# Augmenting emotion features in irony detection with Large language modeling

Yucheng Lin[1], Yuhan Xia[1], Yunfei Long[1]

[1] School of Computer Science and Electronic Engineering, University of Essex, Wivenhoe Park, Colchester CO4 3SQ, UK
`{yl23555, yx23989, yl20051}@essex.ac.uk`

**Abstract.** This study introduces a novel method for irony detection, applying Large Language Models (LLMs) with prompt-based learning to facilitate emotion-centric *text augmentation*. Traditional irony detection techniques typically fall short due to their reliance on static linguistic features and predefined knowledge bases, often overlooking the nuanced emotional dimensions integral to irony. In contrast, our methodology augments the detection process by integrating subtle emotional cues, augmented through LLMs, into three benchmark pre-trained NLP models—BERT, T5, and GPT-2—which are widely recognized as foundational in irony detection. We assessed our method using the SemEval-2018 Task 3 dataset and observed substantial enhancements in irony detection capabilities.

## 1 Introduction

Irony detection, a challenging Natural Language Processing (NLP) task, aims to identify ironic expressions in text. Irony is commonly used in social networks to facilitate social interactions, inspire humour, mitigate or exacerbate criticism, and attract readers' attention through creativity [1]. Research in this area is crucial for improving applications such as emotion analysis, text comprehension, and human-computer interaction, as the presence of irony may affect the true meaning of a text, making automated detection difficult. Irony detection is a complex problem affected by various phenomena, including emotional features, symbols, contextual features, etc. [2].

Researches in irony detection has grown significantly in recent years. Early approaches relied heavily on specific linguistic features and manually labelled datasets [3]. With the development of deep learning techniques, the introduction of the Transformer model in particular has made a huge difference to the NLP field. The model employs a self-attention mechanism that can simultaneously process and analyse all elements in a text sequence, providing contextual relationships throughout the sentence. Its multi-attention architecture enables parallel text analysis at different levels, capturing finer linguistic details [4].

Irony is closely tied to emotion and is often a platform for expressing complex emotional states. It can convey multiple levels of emotion simultaneously, such as entertainment mixed with criticism or feelings mixed with irony. This multilayered expression makes detecting irony particularly challenging, as it requires an understanding of the literal meaning of the words and the emotional intent behind them. Emotions in irony are often expressed subtly and are highly context-dependent, so it is necessary to incorporate irony detection models with complex understanding of emotion [5].

However, existing irony detection methods are limited by their reliance on a fixed knowledge base and predefined linguistic features, making it difficult to adapt to dynamic language and textual diversity [6]. This rigidity leads to their ineffectiveness in identifying and parsing irony in evolving language use, emerging forms of expression, and cross-cultural contexts. In addition, the subjective nature of satire, the two-stage affective process in its interpretation [7], and the influence of non-verbal cues (e.g., emoticons) [8] add to the difficulty of interpreting multi-level affective cues in satire.

The emergence of LLM, such as ChatGPT, has revolutionised progress. These combine large-scale pre-trained language models, contextual learning, and Chained Thinking (CoT). Allowing models to simulate human thought processes and use analogies to learn and complete tasks without relying on manually engineered patterns and features [9] [10]. LLM has shown its unique ability to deal with diverse and complex tasks. Particularly adept at *text augmentation*. This enhances the robustness of NLP models by introducing a broader range of linguistic variations and scenarios, unlike traditional methods that generate datasets less reflective of human language [11]. Additionally, their ability to encode and utilize vast amounts of knowledge as a dynamic, unsupervised knowledge base surpasses traditional knowledge-based approaches [12], making their application more effective and flexible.

Inspired by LLM's application in *text augmentation*. We introduced an innovative method to enhance irony detection by emphasizing emotion features and context features. Initially, we crafted three distinct prompts focusing on emotion, context, and their amalgamation to guide the data generation process. Leveraging the capabilities of GPT-4, we expanded our dataset through these prompts, enriching it with nuanced emotional information. This emotionally augmented dataset was then fed into three sophisticated models, setting the stage for comprehensive training aimed at refining our irony detection methodology. This approach marks a significant advancement in understanding and interpreting the complex interplay of emotion features in textual irony.

The main contributions of this paper are:

-Exploring the effectiveness of *text augmentation* with LLM in irony detection.

-Comparing the effectiveness of different models using different cue words in the prompting process.

-Extensive experiments on the benchmark SemEval-2018 Task 3 dataset prove our model can outperform strong selected baselines by leveraging emotion and context features.

## 2  Related work

The utilization of static external resources to augment the sentiment analysis capabilities of models has been proven effective. For instance, Long et al. introduced an attention model for sentiment analysis, trained using eye-tracking data to enhance word and sentence level focus [13]. Similarly, Xia et al. proposed the incorporation of sensory features of lexical into neural affective analysis frameworks [14]. While effective in standard sentiment analysis tasks, these traditional approaches often struggle with the subtleties of irony that extend beyond mere lexical cues and syntactic structures [15]. Moreover, these methods require extensive extra-linguistic resources and labour-intensive feature engineering, limiting their adaptability [16].

In recent years, various neural network-based models have been proposed and developed for irony detection. For example, Huang et al. investigated how deep learning can be applied to irony detection tasks, primarily through word embedding techniques. They explored deep learning models based on CNN, RNN and RNN with attention mechanism and found that RNN with attention mechanism performs best on the Twitter dataset [17]. Wu et al. proposed a system based on densely connected LSTM and multi-task learning, which achieved excellent results in the SemEval-2018 task [18]. Swanberg et al. Identifying ironic tweets using manually developed features combined with logistic regression models [19]. Van Hee et al. Used SVM combined with a rich feature set for irony detection on manually labelled corpus and compared it with LSTM-based deep learning methods [20].

Despite the tremendous success of the above methods in irony detection, only some utilise external linguistic knowledge to address irony detection. Zhang et al. proposed a transfer learning approach that transfers knowledge from external emotion analysis resources to irony detection, focusing on identifying implicit contradictions without explicit contradiction expressions [21]. Ren et al. explored a knowledge-enhanced neural network model that integrates contextual information from external knowledge sources such as Wikipedia to improve the performance of irony detection [22]. However, these approaches rely on specific knowledge sources and may not adequately capture the implicit complexity of language, especially emotion features. Therefore, we propose using emotion and context knowledge from LLM to bridge this gap and enhance irony detection.

## 3  Methodology

### 3.1 Overall Framework

Our irony detection framework utilizes a two-step approach, combining text enhancement with Large Language Models (LLMs) and advanced text classification methods. In the first step, the text in our dataset is enriched through LLMs. This process deepens the text's complexity, capturing a wide array of contextual and emotional details. This results in a more detailed and expressive text.

After this enhancement, the improved text is analyzed using three different models: T5, GPT-2, and BERT. These models are specialized versions of transformer-based language models known for their top performance in understanding irony. They are designed differently: T5 works as an encoder-decoder, GPT-2 as a decoder, and BERT as an encoder. This variety allows us to thoroughly examine the text, enriched by the LLMs, from multiple perspectives. This detailed examination helps in accurately identifying irony in the text.It's worth noting that our framework is versatile and can be integrated with any classification model. The overarching architecture of our approach is depicted in Figure 1.

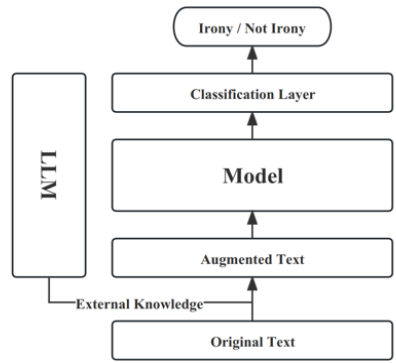

**Fig. 1.** The Architecture of the proposed model

### 3.2 Prompt Based Text Augmentation

For *text augmentation*, we utilised an LLM, specifically GPT-4, to explore emotion features and context features. Previous studies demonstrate that emotion and contextual background play an important role in accurately detecting irony [23] [24]. In cases where ironic content is subtly implied, our framework can reveal its ironic intent more clearly through text augmentation. We designed three of the prompts respectively.

**Emotion Focus Prompt:** *"Expand this sentence to retain the original meaning and expand the emotion words, format: sentence."*

**Contextual Enrichment Prompt:** *"Expand this sentence to retain the original meaning and expand the background of the tweet, format: sentence."*

**Comprehensive Enhancement Prompt:** *"Expand this sentence to retain the original meaning, including elaborating on emotions, and expand the background of the tweet, format: sentence."*

### 3.3 Model Architectures and Classification Methods

In the second step, we used three Transformer models, BERT, T5 and GPT-2, to perform the task of irony detection. The following is a detailed description of these model architectures and their application in the classification process:

BERT uses a multi-layer Transformer encoder architecture where each layer consists of a self-attention mechanism and a feed-forward network. This architecture enables BERT to capture the text's bidirectional relationship by considering each word's contextual information throughout the input sequence [25].

T5 uses an encoder-decoder framework that reframes all NLP tasks as text-to-text transformation problems. The encoder part encodes the input text into a continuous vector representation, while the decoder generates the target text from that representation [26].

GPT-2 is built based on the Transformer decoder architecture, focusing mainly on language generation tasks. Using autoregression, GPT-2 predicts the word at each position, dependent on the words at all previous positions [27].

These models use a similar mechanism in the classification task. Precisely, the classification process consists of the following steps:

Linear transformations:

$$z = W \cdot h + b \qquad (1)$$

Where $h$ represents the model output and $W$ and $b$ are the weights and biases of the linear layer for mapping the model output to the classification space.

Probability Estimation:

$$P(\text{class}) = \sigma(z) \qquad (2)$$

The sigmoid function converts the logits' output from the linear layer into classification probabilities.

## 4 Experiment

In this section, we present the dataset used for irony detection, outline the experimental setup and baseline, and report the results of our proposed method.

### 4.1. Dataset

We selected Task A from the Task 3 dataset of the 2018 Semantic Evaluation (SemEval) workshop for conducting research on irony detection with binary classification. The dataset brings together actual Twitter tweets from different points in time, and the training set contains 3,834 tweets, of which around half are ironic, and half are non-ironic, 1,911 and 1,923, respectively. The test set consists of 311 ironic and 473 non-ironic tweets. This dataset is widely used in irony detection and provides a benchmark for comparison in our study [28].

### 4.2. Baselines

To evaluate our proposed model, we selected the following methods for baseline comparison:

**Van Hee et al. (2018):** The paper focuses on irony detection in English tweets for SemEval 2018 Task 3. The task comprises two subtasks, A and B. Subtask A is a binary classification of whether a tweet expresses irony. Subtask B is a multi-class irony classification for tweets. The paper provides a detailed task description and dataset construction methodology, discussing participants' systems and results. Results show systems performed better on Task A versus B, with the various ironies of comparison easiest to recognise. Additionally, the paper discusses the effects of adding more training data on classification results, suggesting possible explanations [28].

**Zhang et al. (2019):** The research develops three specialised Bi-LSTM models for Twitter irony detection. AABi-LSTM leverages sentiment-focused word corpora, SABi-LSTM uses Twitter sentiment corpora for better contextual understanding, and STBi-LSTM employs a dual Bi-LSTM to enhance sentiment analysis transfer. These models demonstrate superior performance in irony detection by utilising external sentiment knowledge to discern nuanced contexts [29].

**González et al. (2020):** The study introduces a Transformer-based model utilising pre-trained, contextualised Twitter-specific Word2Vec embeddings for irony detection on Twitter. This approach stands out for incorporating contextual word representations, securing top performance in related tasks for Spanish and high performance in English [2].

**Saroj & Pal (2023):** This research compares BERT and ELMo models with the ensemble methods EMLT and EBEM for irony detection in social media texts. Results indicate that BERT's bi-directional context understanding significantly boosts performance, and ensemble methods enhance detection capabilities in this field [30].

**Tasnia et al. (2023):** The study presents a composite neural network model for detecting humour and irony on social media, integrating GloVe, ELMo, BERT, and Flair embeddings with an LSTM layer. The model's diverse contextual representations produce proficient and accurate classification results [31].

### 4.3. Training and Optimization

We standardised the hyperparameters of BERT, T5 and GPT-2 for model training and optimisation, as shown in Table 1. We used the Adam optimiser [32]with a learning rate of 0.00002 and drop out of 0.1 for all models to mitigate overfitting. While the BERT model was trained for 7 epochs, both T5 and GPT-2 underwent 10 epochs of training, balancing robustness in irony detection with the avoidance of overfitting. Our algorithm setup table outlines the strategy and aims to optimise model performance in terms of efficiency and accuracy.

| Model | Optimizer | Learning Rate | Dropout | Epoch | Model Variant |
|---|---|---|---|---|---|
| Bert | Adam | 0.00002 | 0.1 | 7 | Bert-Base-Uncased |
| T5 | Adam | 0.00002 | 0.1 | 10 | T5-Base |
| GPT | Adam | 0.00002 | 0.1 | 10 | GPT2 |

**Table 1.** Algorithms settings

| Prompt | Evaluation Metrics(%) | | | | | | | | | | | |
|---|---|---|---|---|---|---|---|---|---|---|---|---|
| | BERT-base | | | | T5-base | | | | GPT2 | | | |
| | Precision | Recall | $F_1$ | Acc | Precision | Recall | $F_1$ | Acc | Precision | Recall | $F_1$ | Acc |
| No prompt | 64.5 | 85.8 | 69.2 | 72.3 | 68.2 | **88.4** | 73.1 | 73.7 | 60.3 | 70.6 | 62.6 | 68.0 |
| Emotion Focus Prompt | 79.4 | 72.7 | 72.3 | 78.4 | **76.0** | 71.2 | 68.7 | 75.3 | 63.8 | **81.6** | **66.7** | 69.8 |
| Contextual Enrichment Prompt | **79.6** | 76.3 | 73.4 | 77.8 | 64.1 | 69.6 | 64.6 | 70.9 | 64.6 | 76.6 | 63.9 | 69.4 |
| Comprehensive Enhancement Prompt | 76.9 | **89.2** | **78.2** | **81.8** | 72.0 | 82.8 | **73.4** | **77.6** | **66.2** | 79.6 | 66.6 | **72.1** |

**Table 2.** Best results obtained in irony detection

| Model | Precision | Recall | $F_1$ | Acc |
|---|---|---|---|---|
| Van Hee et al. (2018)* | **78.8** | 66.9 | 72.4 | 79.7 |
| Zhang et al. (2019) | - | - | 73.6 | - |
| González et al. (2020) | 57.2 | 84.6 | 68.2 | 68.9 |
| Saroj & Pal (2023) | 64.5 | 70.2 | 67.2 | 67.9 |
| Tasnia et al. (2023) | 59.3 | **89.4** | 71.2 | 71.4 |
| **Our Model** | 76.9 | 89.2 | **78.2** | **81.8** |

**Table 3.** Model performance rankings for semeval 2018 Task A. * indicates the top-performing team in this study.

### 4.4. Headline results evaluation

In our study, performance is evaluated using established metrics, namely accuracy, precision, recall, and the F1 score. To ensure the reliability and consistency of our evaluation, we employ a methodical approach. The model is trained in several iterations with stable hyperparameters, and the final evaluation metrics are determined by averaging the best-performing ephemeral elements from each iteration.

The empirical data, detailed in Table 2, showcase consistent enhancements across various models upon the integration of targeted cues for irony detection. Specifically, in the BERT-base model, the incorporation of cues that include extended emotional vocabulary boosts the accuracy from a baseline of 72.3% to 78.4%. Moreover, the further addition of contextual background information escalates the accuracy to 81.8%.

Meanwhile, additional context information, noted as expand the background, also significantly enhances the performance of NLP models, especially in the case of the BERT-base model. With this expanded context, the model demonstrates a notable increase in accuracy by 5.5%, but recall decreases by 9.5%. However, with the combined application of emotion analysis and expanded context, the BERT-base model experiences an increase in recall alongside a total accuracy gain of 9.5% and an improvement in recall of 3.4%. Although performance gains on T5-base and GPT-2 models are relatively modest, they still benefit from different *text augmentation* strategies, suggesting effectiveness in improving overall model performance on specific evaluation metrics. This trend emphasises the critical role of comprehensive contextual analysis in enhancing the accuracy of irony detection across various NLP models.

Table 3 delineates the model rankings based on their performance in the irony detection task of SemEval 2018 Task A, quantified by F1 scores. Notably, our model achieves an F1 score of 78.2%, thereby surpassing the top-performing team result in Van Hee et al. (2018) as well as our own baseline model. This underscores significant advancements in the domain of irony detection.

**4.6. Case Analysis**

In irony detection, contextual and emotional information is crucial. For this reason, we will present a case study by comparing the original text with the text enhanced by emotion and context. Our case is shown in Fig 2.

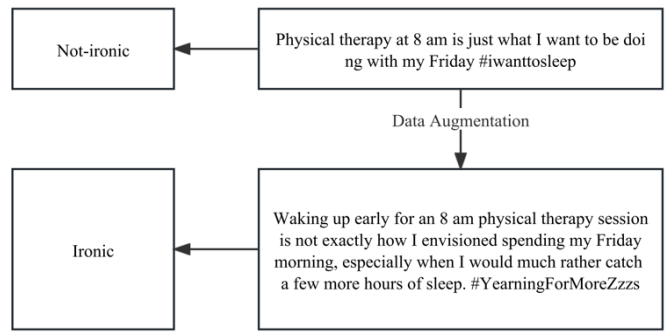

**Fig. 2.** Case study example figure, the top is the original text, the bottom is the text after data enhancement.

In this example, the irony of the original sentence is reflected in the apparent anticipation of a morning physiotherapy session, but it is a resentment of this and a desire for extra sleep time, a direct reference to a morning schedule, but the irony is not apparent. Without *text augmentation* using LLM for emotion and context, the model may have interpreted the sentence as a simple statement and ignored the true feelings about the situation. When additional information is added, the *augmented text* reflects a contrast between the expectation of an ideal state of affairs in the morning and the reality. This contrast makes the irony more pronounced.

# 5 Conclusion

This study explores integrating a large language model (LLM) for irony detection via text augmentation. The approach involves extracting emotion and contextual features from text using GPT-4 to enrich the data before inputting it into machine learning models such as BERT, T5, and GPT-2. The experimental results, benchmarked against the SemEval-2018 Task 3 dataset, demonstrate that this approach outperforms existing models. This study shows that LLM has excellent potential to enhance the performance of NLP tasks that require a deeper understanding of language.

For future research, the focus will be on exploring LLM potential in processing and optimising irony detection. We plan to investigate the impact of LLM when pre-processing text, focusing on elements such as emoticons and personal names that are often rich in emotion and complex context. This series of studies aims to enhance our comprehension of how these elements influence the interpretation of irony, particularly in informal and diverse communication channels such as social media. Furthermore, we intend to expand the LLM-based approach to multiple languages, evaluating its effectiveness and adaptability in various linguistic contexts.

**Acknowledgments.** This work is supported by the Alan Turning Institute/DSO grant: Improving multimodality misinformation detection with affective analysis . Yunfei Long acknowledges the financial support of the School of Computer science and Electrical Engineering, University of Essex.